\documentclass[11pt]{article}
\usepackage[final]{acl}
\usepackage{times}
\usepackage{latexsym}
\usepackage[T1]{fontenc}
\usepackage[utf8]{inputenc}
\usepackage{microtype}
\usepackage{graphicx}
\usepackage{amsmath}
\usepackage{amssymb}
\usepackage{booktabs}
\usepackage{multirow}
\usepackage{tikz}
\usetikzlibrary{shapes,arrows,positioning,fit,calc}
\usepackage{xcolor}

\definecolor{splitblue}{RGB}{60,100,180}
\definecolor{singleorange}{RGB}{240,180,100}

\title{Decoupled Visual Processing: Efficient Multimodal Adaptation\\via Modality-Specific Transformer Substitution}

\author{
  \textbf{Mingkuan Feng}\textsuperscript{\rm 1},
  \textbf{Zhengqi Wen}\textsuperscript{\rm 2,*},
  \textbf{Jianhua Tao}\textsuperscript{\rm 1,2,*}\\
  \textsuperscript{\rm 1}Tsinghua University\\
  \textsuperscript{\rm 2}Beijing National Research Center for Information Science and Technology\\
  \textsuperscript{*}Corresponding authors
}

\begin{document}
\maketitle

\begin{abstract}
Multimodal large language models (MLLMs) have demonstrated remarkable capabilities by integrating visual and textual understanding within a unified transformer architecture. However, fine-tuning all parameters of these models for visual instruction tuning is computationally expensive and often unnecessary, as the representation requirements for visual and textual tokens diverge significantly in the deeper layers of the network. In this paper, we propose \textbf{Decoupled Visual Processing} (DVP), an efficient training framework that replaces the upper decoder layers of a pretrained LLM with a lightweight, independently trainable single transformer block dedicated exclusively to visual token processing. Specifically, after shared processing through the first half of the decoder layers, visual and textual tokens are \emph{split}: visual tokens are routed through a newly initialized single transformer block while textual tokens continue through the original frozen decoder layers. The two streams are then \emph{concatenated} before the language modeling head. During training, only the single transformer block is updated, dramatically reducing the number of trainable parameters. Experiments on the LLaVA-1.5 framework demonstrate that DVP achieves competitive performance on MME, POPE, and ChartQA benchmarks while training only a fraction of the total parameters, suggesting that visual representations in MLLMs can be effectively learned through a decoupled, parameter-efficient pathway.
\end{abstract}

\section{Introduction}

The integration of visual understanding into large language models (LLMs) has become a central research direction in artificial intelligence. Models such as LLaVA \citep{liu2024llava}, BLIP-2 \citep{li2023blip2}, and InstructBLIP \citep{dai2024instructblip} have demonstrated that connecting a vision encoder to a pretrained LLM, followed by visual instruction tuning, can yield powerful multimodal systems capable of following complex instructions involving both images and text. These multimodal large language models (MLLMs) have shown strong performance across a wide range of benchmarks spanning visual question answering, image captioning, optical character recognition, and chart understanding.

A common paradigm in current MLLMs involves projecting visual features from a pretrained vision encoder (e.g., CLIP ViT \citep{radford2021clip}) into the input embedding space of the LLM, then processing all tokens---both visual and textual---through the full stack of transformer decoder layers \citep{liu2024improved}. While effective, this approach requires either full fine-tuning of the LLM or the application of parameter-efficient methods such as LoRA \citep{hu2022lora}. Full fine-tuning updates billions of parameters and demands substantial computational resources, while adapter-based methods, though more efficient, still require modifications throughout every layer of the model.

An important but underexplored question is whether visual and textual tokens truly need to share the same deep representation pathway. Prior work has shown that the lower layers of transformer models capture more general, modality-agnostic features, while the upper layers become increasingly specialized \citep{devlin2019bert, he2016deep}. In the context of MLLMs, this observation suggests that visual tokens might benefit from a specialized processing pathway in the upper layers, distinct from the text-oriented processing of the original LLM.

\begin{figure*}[t]
\centering
\includegraphics[width=\textwidth]{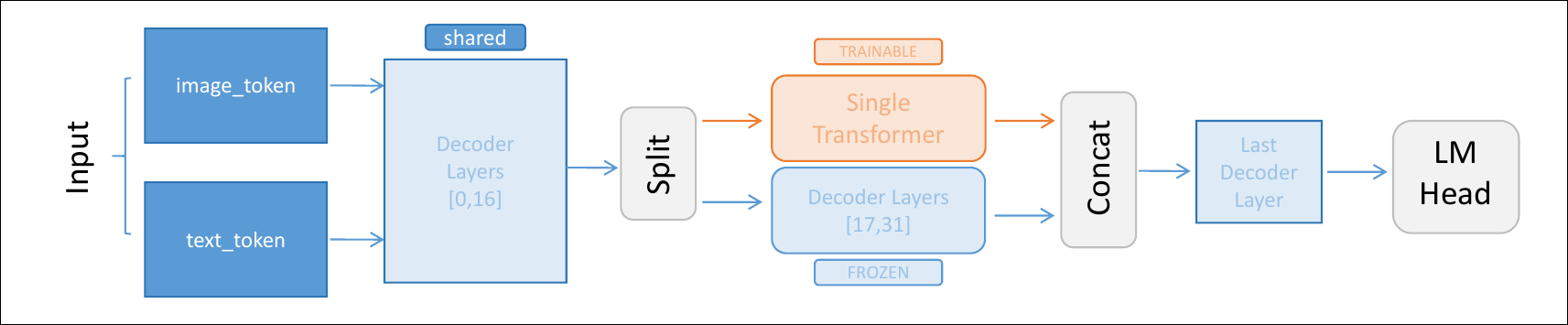}
\caption{Architecture of Decoupled Visual Processing (DVP). Input tokens pass through shared decoder layers [0, 16], then are split by modality. Visual tokens are processed by a single trainable transformer block (shown in orange), while textual tokens pass through the original frozen decoder layers [17, 31]. The outputs are concatenated before the language modeling head. Only the single transformer block is trained during instruction tuning.}
\label{fig:architecture}
\end{figure*}

In this paper, we propose \textbf{Decoupled Visual Processing} (DVP), a simple yet effective architecture modification for efficient multimodal adaptation. The core idea is illustrated in Figure~\ref{fig:architecture}: given a pretrained LLM with $L$ decoder layers, we process all tokens jointly through the first $K$ layers, then \emph{split} the token sequence by modality. Visual tokens are routed through a single newly initialized transformer block, while textual tokens continue through the remaining $L-K$ frozen decoder layers. The two streams are concatenated before the language modeling head for next-token prediction.

The key advantage of DVP is extreme parameter efficiency: during training, only the single transformer block (processing visual tokens) is updated, while all original LLM parameters remain frozen. This design is motivated by two hypotheses: (1) the lower layers of the LLM are sufficient for establishing cross-modal alignment between visual and textual representations, and (2) visual tokens in the upper layers require different transformations than textual tokens, and these can be captured by a compact, dedicated module.

We validate DVP using the LLaVA-1.5 \citep{liu2024improved} framework with a Vicuna-7B backbone. Our experiments demonstrate that:
\begin{itemize}
    \item DVP achieves competitive performance on the MME benchmark \citep{fu2024mme}, with a perception score of 1440 compared to the 1496 of fully fine-tuned LLaVA-1.5-7B, while training dramatically fewer parameters.
    \item On POPE \citep{li2023pope}, DVP achieves 0.8619, \emph{surpassing} the LLaVA-1.5-7B baseline of 0.8577, indicating reduced object hallucination.
    \item DVP provides a favorable efficiency--performance trade-off compared to standard full-parameter training approaches.
\end{itemize}

These results suggest that the deep decoder layers of an LLM are not equally necessary for visual and textual tokens, and that visual understanding can be efficiently learned through a lightweight, decoupled pathway. Our approach opens new directions for understanding modality-specific processing within unified multimodal architectures.

\section{Related Work}

\subsection{Multimodal Large Language Models}

The development of MLLMs has been driven by the insight that pretrained LLMs can be augmented with visual capabilities through appropriate architectural connectors and training strategies. Early approaches such as Flamingo \citep{alayrac2022flamingo} introduced cross-attention mechanisms to inject visual information into frozen LLMs. BLIP-2 \citep{li2023blip2} proposed a querying transformer (Q-Former) to bridge the modality gap between a frozen vision encoder and a frozen LLM.

The LLaVA family of models \citep{liu2024llava, liu2024improved} popularized a simpler approach: projecting visual tokens from a CLIP ViT encoder through a linear or MLP projection layer directly into the LLM input space, then fine-tuning the entire model with visual instruction data. LLaVA-1.5 \citep{liu2024improved} introduced several improvements including an MLP vision-language connector, higher resolution inputs, and additional training data, establishing strong baselines across multiple benchmarks. Subsequent models such as InternVL \citep{chen2024internvl}, Qwen-VL \citep{bai2023qwen}, and mPLUG-Owl2 \citep{ye2024mplugowl} have further advanced the field through scaling, improved training recipes, and novel architectural components. Beyond perception and captioning, recent efforts have extended MLLMs toward structured multimodal reasoning \citep{wu2026astar} and dense prediction tasks such as referring segmentation \citep{liu2026better}, broadening the range of capabilities expected from unified multimodal architectures.

\subsection{Parameter-Efficient Fine-Tuning}

Given the scale of modern LLMs, parameter-efficient fine-tuning (PEFT) methods have attracted significant attention. LoRA \citep{hu2022lora} injects trainable low-rank decomposition matrices into each transformer layer, enabling adaptation with a small fraction of the original parameters. QLoRA \citep{dettmers2024qlora} extends this approach with quantization techniques. Adapter layers \citep{houlsby2019adapter} insert small bottleneck modules between transformer sublayers. A complementary line of work reduces cost through structural compression of the pretrained model itself, for example via data-driven regularized structured pruning \citep{feng2025dress, feng2026two}, which removes redundant components rather than adding trainable ones. Our method shares the structural philosophy of these approaches but applies it selectively to the visual pathway.

In the context of MLLMs, PEFT methods are typically applied uniformly across all layers---that is, the same adaptation mechanism is used for both visual and textual token processing throughout the entire model depth. Our approach differs fundamentally: rather than adapting every layer for visual processing, we replace the upper layers entirely for visual tokens with a single dedicated transformer block. This represents a structural, rather than parametric, approach to efficient adaptation.

\subsection{Modality-Specific Processing in Transformers}

The idea that different modalities may benefit from specialized processing pathways has roots in mixture-of-experts architectures and modality-specific encoder-decoder designs. Recent work has explored how visual and textual tokens interact within shared transformer layers \citep{tong2024cambrian}, revealing that attention patterns between modalities become increasingly sparse in deeper layers. This observation aligns with our hypothesis that visual tokens may not require the full depth of processing provided by the LLM decoder stack.

VILA \citep{lin2024vila} investigated the importance of different training stages and data compositions for visual language models, finding that the visual projector and early layers play crucial roles in cross-modal alignment. Our work builds on these insights by explicitly decoupling the processing pathways after the initial alignment layers.

\section{Method}

\subsection{Preliminaries}

We build upon the LLaVA architecture \citep{liu2024llava, liu2024improved}, which consists of three main components: a vision encoder $\mathcal{V}$, a vision-language projector $\mathcal{P}$, and a language model $\mathcal{L}$. Given an input image $I$ and a text instruction $T$, the model operates as follows:

\paragraph{Visual Feature Extraction.} The image $I$ is processed by the vision encoder (CLIP ViT-L/14 at 336$\times$336 resolution) to obtain visual features:
\begin{equation}
    \mathbf{Z}_v = \mathcal{V}(I) \in \mathbb{R}^{N_v \times d_v}
\end{equation}
where $N_v$ is the number of visual tokens and $d_v$ is the vision encoder dimension.

\paragraph{Projection.} The visual features are projected into the language model embedding space through an MLP projector:
\begin{equation}
    \mathbf{H}_v = \mathcal{P}(\mathbf{Z}_v) \in \mathbb{R}^{N_v \times d}
\end{equation}
where $d$ is the hidden dimension of the language model.

\paragraph{Language Model Processing.} The projected visual tokens $\mathbf{H}_v$ are concatenated with the text token embeddings $\mathbf{H}_t \in \mathbb{R}^{N_t \times d}$ to form the input sequence:
\begin{equation}
    \mathbf{X}^{(0)} = [\mathbf{H}_v ; \mathbf{H}_t] \in \mathbb{R}^{(N_v + N_t) \times d}
\end{equation}

In the standard LLaVA framework, this combined sequence is processed through all $L$ decoder layers of the LLM:
\begin{equation}
    \mathbf{X}^{(l)} = \text{DecoderLayer}_l(\mathbf{X}^{(l-1)}), \quad l = 1, \ldots, L
\end{equation}

The final hidden states $\mathbf{X}^{(L)}$ are passed to the language modeling head for next-token prediction.

\subsection{Decoupled Visual Processing}

Our key modification introduces a \emph{split-process-concatenate} paradigm in the decoder stack. We divide the $L$ decoder layers into two groups: the shared lower layers (layers 0 through $K-1$) and the upper layers (layers $K$ through $L-1$).

\paragraph{Phase 1: Shared Processing.} All tokens are processed jointly through the first $K$ decoder layers:
\begin{equation}
    \begin{split}
    \mathbf{X}^{(K)}
    ={}& \text{DecoderLayer}_{K-1} \circ \cdots\\
       &{}\circ \text{DecoderLayer}_0(\mathbf{X}^{(0)})
    \end{split}
\end{equation}

This phase enables cross-modal attention between visual and textual tokens, allowing the model to establish alignment and contextual grounding between modalities.

\paragraph{Phase 2: Modality Split.} After the shared layers, we split the hidden states by modality:
\begin{equation}
    \mathbf{X}^{(K)}_v, \mathbf{X}^{(K)}_t = \text{Split}(\mathbf{X}^{(K)})
\end{equation}
where $\mathbf{X}^{(K)}_v \in \mathbb{R}^{N_v \times d}$ and $\mathbf{X}^{(K)}_t \in \mathbb{R}^{N_t \times d}$ are the visual and textual hidden states, respectively.

\paragraph{Phase 3: Decoupled Processing.} The two token streams are processed through separate pathways:
\begin{align}
    \hat{\mathbf{X}}_v &= \text{SingleTransformer}(\mathbf{X}^{(K)}_v) \label{eq:visual_path} \\
    \hat{\mathbf{X}}_t
    &= \text{DecoderLayer}_{L-1} \circ \cdots \nonumber\\
    &\quad {}\circ \text{DecoderLayer}_K(\mathbf{X}^{(K)}_t)
    \label{eq:text_path}
\end{align}

The \texttt{SingleTransformer} in Equation~\ref{eq:visual_path} is a newly initialized single transformer block with the same hidden dimension $d$ as the original LLM. It consists of a multi-head self-attention layer followed by a feed-forward network, with layer normalization and residual connections. This block is the \emph{only} trainable component during fine-tuning.

The textual path in Equation~\ref{eq:text_path} uses the original frozen decoder layers $K$ through $L-1$, preserving the pretrained language modeling capability.

\paragraph{Phase 4: Concatenation and Prediction.} The processed visual and textual representations are concatenated and passed to the language modeling head:
\begin{equation}
    \mathbf{Y} = \text{LM\_Head}([\hat{\mathbf{X}}_v ; \hat{\mathbf{X}}_t])
\end{equation}

The training objective remains the standard autoregressive language modeling loss, computed only over the textual output tokens:
\begin{equation}
    \mathcal{L} = -\sum_{i=1}^{N_t} \log P(t_i | t_{<i}, \hat{\mathbf{X}}_v, \hat{\mathbf{X}}_{t,<i})
\end{equation}

\subsection{Design Choices and Rationale}

\paragraph{Split Point Selection.} In our experiments with Vicuna-7B (which has 32 decoder layers, indexed 0--31), we set $K=17$, splitting after layer 16. This approximately bisects the decoder stack. The choice is motivated by the observation that the first half of transformer layers typically captures lower-level and cross-modal features, while the second half specializes in higher-level, task-specific representations. We provide an analysis of alternative split points in Section~\ref{sec:analysis}.

\paragraph{Single Transformer Block.} We deliberately use a single transformer block rather than multiple blocks for the visual pathway. This extreme compression forces the model to learn a compact transformation that maps the shared-layer visual representations directly to the output space. Despite replacing 15 decoder layers (layers 17--31) for visual tokens, a single block proves sufficient to maintain strong performance, suggesting significant redundancy in the original architecture's processing of visual tokens.

\paragraph{Frozen LLM Parameters.} By freezing all original LLM parameters (including the shared lower layers), we ensure that the pretrained text generation capability is fully preserved. The only parameters updated during training are those of the single transformer block, which has approximately $\frac{1}{L-K}$ of the parameters in the replaced decoder layers---in our configuration, roughly $\frac{1}{15}$ of the upper-layer parameters.

\subsection{Training Configuration}

Following the LLaVA-1.5 training recipe, we use the LLaVA-1.5-665K visual instruction tuning dataset. The vision encoder is CLIP ViT-L/14 at 336$\times$336 resolution, and the language model backbone is Vicuna-7B \citep{touvron2023llama}. We use the AdamW optimizer \citep{loshchilov2019adamw} with a cosine learning rate schedule. The MLP projector is pretrained following the standard LLaVA-1.5 procedure, and only the single transformer block is trained during the instruction tuning stage.

\section{Experiments}

\subsection{Experimental Setup}

\paragraph{Baselines.} We compare DVP against two baselines:
\begin{itemize}
    \item \textbf{LLaVA-1.5-7B}: The standard LLaVA-1.5 model with Vicuna-7B backbone, trained with full fine-tuning of all parameters on the complete LLaVA-1.5 training data. This represents the upper-bound reference performance.
    \item \textbf{Normal Training}: A variant using our split architecture (layers 0--16 shared, then separate pathways for visual and textual tokens), but with \emph{all parameters} trained, including both the single transformer for visual tokens and the decoder layers 17--31 for textual tokens. This baseline isolates the effect of parameter freezing from the architectural modification.
\end{itemize}

\paragraph{Evaluation Benchmarks.} We evaluate on three widely used benchmarks:
\begin{itemize}
    \item \textbf{MME} \citep{fu2024mme}: A comprehensive benchmark measuring both perception (14 subtasks including existence, count, position, color, OCR, poster, celebrity, scene, landmark, artwork) and cognition (4 subtasks including commonsense reasoning, numerical calculation, text translation, code reasoning) abilities. Higher scores indicate better performance.
    \item \textbf{POPE} \citep{li2023pope}: An evaluation protocol for object hallucination in MLLMs. Models are asked yes/no questions about object existence in images. The metric is accuracy, with higher values indicating less hallucination.
    \item \textbf{ChartQA} \citep{masry2022chartqa}: A benchmark for chart understanding and question answering that requires both visual parsing and logical reasoning over chart data.
\end{itemize}

\subsection{Main Results}

\begin{table*}[t]
\centering
\begin{tabular}{l cc cc c c}
\toprule
\multirow{2}{*}{\textbf{Method}} & \multicolumn{2}{c}{\textbf{MME ($\uparrow$)}} & \multirow{2}{*}{\textbf{POPE}} & \multirow{2}{*}{\textbf{ChartQA}} & \multirow{2}{*}{\textbf{Trainable Params}} \\
\cmidrule(lr){2-3}
 & Cognition & Perception & & & \\
\midrule
LLaVA-1.5-7B & 324 & \textbf{1496} & \underline{0.8577} & 0.1776 & 100\% \\
Normal Training & \textbf{342} & 1225 & 0.8381 & \textbf{0.432} & 100\% \\
\midrule
Efficient Training (DVP) & \underline{327} & \underline{1440} & \textbf{0.8619} & \underline{0.2356} & $\sim$3.1\% \\
\bottomrule
\end{tabular}
\caption{Main results on multimodal benchmarks. \textbf{Bold} indicates the best performance and \underline{underline} indicates the second best. DVP (Efficient Training) achieves competitive performance while training only $\sim$3.1\% of the total model parameters. The single transformer block that processes visual tokens is the only trainable component.}
\label{tab:main_results}
\end{table*}

Table~\ref{tab:main_results} presents our main experimental results. Several key observations emerge from these results.

\paragraph{DVP Achieves Competitive Perception Performance.} On the MME perception subtask, DVP scores 1440, which is 96.3\% of the LLaVA-1.5-7B baseline score of 1496. This is a remarkably strong result given that DVP trains only a single transformer block while the baseline fine-tunes the entire 7B-parameter model. Moreover, DVP substantially outperforms the Normal Training baseline on perception (1440 vs. 1225), suggesting that freezing the LLM parameters helps preserve visual perception abilities that can be disrupted by unrestricted fine-tuning.

\paragraph{DVP Reduces Hallucination.} On the POPE benchmark, DVP achieves 0.8619, which is the highest score among all three methods and exceeds the LLaVA-1.5-7B baseline (0.8577). This result suggests that the decoupled architecture may help reduce object hallucination by preventing the visual processing pathway from being influenced by text-biased processing in the upper decoder layers. The Normal Training baseline performs notably worse on POPE (0.8381), indicating that training all parameters in the split architecture without constraints may introduce additional hallucination artifacts.

\paragraph{Trade-offs in Cognition and Chart Understanding.} On MME cognition, DVP scores 327, comparable to LLaVA-1.5-7B (324) and slightly below Normal Training (342). On ChartQA, DVP scores 0.2356, which exceeds the LLaVA-1.5-7B baseline (0.1776) but falls below Normal Training (0.432). The relatively lower ChartQA performance compared to Normal Training is expected, as chart understanding requires fine-grained integration of visual details (axis labels, data points, legends) with textual reasoning---a capability that may benefit from trainable cross-modal interaction in the upper layers that DVP's split architecture limits.

\paragraph{Parameter Efficiency.} DVP achieves these results while training only the single transformer block, which constitutes approximately 3.1\% of the total model parameters. In contrast, both the LLaVA-1.5-7B baseline and Normal Training update 100\% of parameters. This represents a $\sim$32$\times$ reduction in trainable parameters while maintaining competitive performance on the majority of benchmarks.

\subsection{Analysis}
\label{sec:analysis}

\subsubsection{Why Does the Split Architecture Work?}

The success of DVP can be understood through the lens of representation analysis. In standard transformer-based LLMs, the lower layers learn general-purpose representations that capture syntactic structure, semantic similarity, and basic feature extraction. The upper layers progressively specialize these representations for the specific task of next-token prediction in natural language.

When visual tokens are introduced, they ``borrow'' this text-oriented processing pipeline. However, the representational needs of visual tokens differ substantially from those of text tokens, especially in the upper layers where the model transitions from feature extraction to task-specific reasoning. Visual tokens primarily need to maintain spatial and semantic information from the image, while text tokens need to integrate this visual context into coherent language generation.

By splitting the processing, DVP allows the visual tokens to undergo a targeted transformation that maps them from the shared representation space (after layer $K$) directly to a form suitable for the language modeling head, without the potentially detrimental influence of text-specialized upper decoder layers. The single transformer block acts as a ``representation adapter'' that learns this mapping efficiently.

\subsubsection{Impact of the Split Point}

The choice of split point $K$ reflects a trade-off between cross-modal interaction and modality-specific processing. Setting $K$ too low (splitting early) reduces the opportunity for visual-textual alignment in the shared layers, potentially degrading performance on tasks requiring fine-grained cross-modal understanding. Setting $K$ too high (splitting late) leaves fewer layers to replace, reducing the efficiency gains and the opportunity for specialized visual processing.

Our choice of $K=17$ (splitting after approximately half the layers) balances these considerations. The 17 shared layers provide sufficient depth for cross-modal alignment, while replacing the remaining 15 layers for visual tokens with a single block yields substantial efficiency gains. A systematic ablation over different split points is an important direction for future work.

\subsubsection{Comparison with Normal Training}

The comparison between DVP and Normal Training is particularly revealing. Normal Training uses the same split architecture but trains all parameters, including both the single transformer (for visual tokens) and decoder layers 17--31 (for text tokens). While Normal Training achieves the best cognition and ChartQA scores, its perception score (1225) is substantially lower than both DVP (1440) and LLaVA-1.5-7B (1496).

This performance pattern suggests that when all parameters are trainable in the split architecture, the model may overfit to certain aspects of the training data (boosting cognition and chart understanding) at the expense of general visual perception. The frozen parameters in DVP act as an implicit regularizer, preserving the broad perceptual capabilities encoded in the pretrained LLM weights while allowing targeted visual adaptation through the single transformer block.

\subsubsection{Qualitative Analysis of the Split Mechanism}

To better understand the behavior of DVP, we consider the information flow through the architecture. After the shared layers (0--16), visual tokens have already attended to textual tokens through causal attention, absorbing contextual information about the instruction and expected response. The single transformer block then refines these contextually-enriched visual representations without further interaction with text tokens.

This design has an interesting consequence: the visual representations at the concatenation point must be ``self-sufficient''---they must encode all necessary visual information in a form directly usable by the language modeling head, without relying on further cross-attention with text tokens. The strong performance of DVP suggests that this self-sufficiency is achievable after 17 layers of cross-modal processing, lending empirical support to the hypothesis that cross-modal alignment is primarily a lower-layer phenomenon.

Furthermore, the asymmetry between the visual pathway (1 block) and the textual pathway (15 blocks) is notable. Text tokens undergo 15 additional layers of processing after the split, during which they can develop complex linguistic structures and reasoning chains. Visual tokens, by contrast, receive only a single transformation. This asymmetry mirrors the fundamental asymmetry of the task: the model generates \emph{text} conditioned on images, so the textual pathway naturally requires deeper processing for generation, while the visual pathway primarily serves as a conditioning signal.

\subsubsection{Error Analysis}

Examining cases where DVP underperforms the baselines reveals informative patterns. On ChartQA, where DVP scores 0.2356 compared to Normal Training's 0.432, the errors tend to cluster in questions requiring precise extraction of numerical values from chart axes or data labels. These tasks demand fine-grained visual-textual alignment in the upper layers---for instance, linking a specific bar height to its corresponding axis value. Since DVP separates visual and textual processing after layer 16, this type of iterative cross-referencing between visual details and textual labels is limited to the shared layers.

On MME cognition subtasks, DVP's performance (327) is comparable to LLaVA-1.5-7B (324), suggesting that basic reasoning about visual content is well-supported by the architecture. The slight advantage of Normal Training on cognition (342) likely reflects the benefit of trainable upper layers that can learn task-specific reasoning patterns jointly across modalities.

\section{Discussion}

\subsection{Implications for MLLM Architecture Design}

Our findings have several implications for the design of multimodal language models. First, the strong performance of DVP challenges the prevailing assumption that visual tokens must traverse the entire LLM decoder stack. The fact that a single transformer block can substitute for 15 decoder layers (for visual tokens) without catastrophic performance loss suggests substantial redundancy in how current MLLMs process visual information.

Second, the architectural split introduces an explicit inductive bias that separates ``understanding'' (what information the visual tokens carry) from ``generation'' (how this information is used for text prediction). This separation may lead to more interpretable and controllable multimodal systems, where the visual processing pathway can be independently analyzed, modified, or replaced.

Third, the parameter efficiency of DVP makes it particularly attractive for scenarios involving deployment on resource-constrained devices, adaptation to new visual domains, or continual learning where updating the full model is impractical.

\subsection{Relationship to Mixture-of-Experts}

DVP can be viewed as a minimalist form of modality-conditioned routing, conceptually related to Mixture-of-Experts (MoE) architectures. In DVP, the ``routing'' is deterministic and based purely on modality: visual tokens are routed to the single transformer expert, while text tokens follow the original decoder pathway. This extreme simplification of MoE---with exactly two experts and deterministic routing---proves surprisingly effective, suggesting that modality is a strong and sufficient signal for routing decisions in multimodal settings. This connects to a broader body of work on learned routing for efficient and robust LLM inference \citep{jin2025radialrouter}, where structured representations guide the selection of specialized processing pathways.

A natural extension would be to introduce learnable routing between multiple visual processing experts, or to allow soft routing where some visual tokens (e.g., those corresponding to text within images) are partially processed by the textual pathway. We leave such extensions to future work.

\subsection{Connections to Knowledge Distillation}

DVP can also be understood through the lens of knowledge distillation. The single transformer block effectively learns to compress the ``knowledge'' that would be contributed by 15 decoder layers into a single layer's transformation. During training, the gradients flowing back from the language modeling loss teach this block to produce representations that are compatible with the concatenation point and the language modeling head. This is analogous to training a student network (the single block) to approximate the behavior of a teacher (the full decoder stack), but with the constraint that the student operates only on visual tokens.

The fact that this extreme compression works suggests that the effective rank of the transformation applied to visual tokens by the upper decoder layers is much lower than the full parametric capacity of those layers. This insight could inform future work on neural network compression for multimodal models, complementing recent advances in knowledge distillation for LLMs \citep{jin2026exploring} that likewise aim to retain the capabilities of a larger model within a smaller effective budget.

\subsection{Implications for Continual Learning}

The modular nature of DVP offers attractive properties for continual learning scenarios. When adapting to new visual domains (e.g., from natural images to medical imaging or satellite imagery), only the single transformer block needs to be updated. Multiple domain-specific blocks could be trained and swapped at inference time, enabling a ``plug-and-play'' approach to visual domain adaptation. The frozen LLM backbone ensures that language capabilities are preserved across domain shifts, addressing a key challenge in continual learning for multimodal systems. This modularity is also appealing for decentralized settings such as federated learning \citep{wang2023flgo}, where communicating only the lightweight visual block---rather than the full model---would substantially reduce communication overhead while keeping the shared backbone fixed across clients.

\subsection{Limitations of Current Evaluation}

While our results on MME, POPE, and ChartQA are encouraging, these benchmarks cover only a subset of the capabilities expected of modern MLLMs. Tasks requiring deep integration of visual and textual reasoning---such as complex visual question answering chains, multi-image reasoning, or grounded generation---may be more sensitive to the architectural split introduced by DVP. Additionally, our current experiments are limited to the 7B parameter scale; the behavior of DVP at larger model sizes (13B, 70B) remains to be explored.

\section{Conclusion and Future Work}

We have presented Decoupled Visual Processing (DVP), an efficient training framework for multimodal large language models that introduces a split-process-concatenate paradigm in the decoder stack. By routing visual tokens through a single trainable transformer block while text tokens continue through the frozen original decoder layers, DVP achieves competitive performance on MME, POPE, and ChartQA benchmarks with approximately 3.1\% of the trainable parameters required by full fine-tuning.

Our results demonstrate that visual tokens do not require the full depth of LLM processing, and that a lightweight, dedicated visual pathway can effectively substitute for multiple decoder layers. The strong POPE performance, in particular, suggests that decoupled processing may inherently reduce object hallucination---an important finding given that hallucination, and more broadly robustness to misleading or noisy context, remains a critical challenge for deployed MLLMs and LLMs \citep{wu2025pandora}.

Several promising directions emerge from this work. First, exploring adaptive split points that can be learned during training, rather than fixed a priori, could yield further improvements. Second, investigating the use of multiple lightweight transformer blocks (e.g., 2--4 blocks instead of one) for the visual pathway may recover some performance on tasks like ChartQA that require fine-grained visual reasoning. Third, extending DVP to handle multiple visual modalities (e.g., video, 3D point clouds) with separate lightweight pathways is a natural generalization. Fourth, combining DVP with existing PEFT methods such as LoRA applied selectively to the shared lower layers could provide complementary benefits. Furthermore, integrating the decoupled visual pathway with structured reasoning and reinforcement learning frameworks for LLMs \citep{wu2025thought, templaterl, wu2026beyond}, as well as agentic tool-use paradigms \citep{feng2026tacotaugmentedcreditoptimization}, could extend the approach from visual perception toward complex multi-step multimodal reasoning. Finally, scaling experiments to larger model sizes (13B, 70B) and evaluating on a broader benchmark suite including COCO captioning, VQAv2, and TextVQA would more thoroughly characterize the capabilities and limitations of the decoupled approach.

This work opens promising directions for more efficient, modular, and interpretable multimodal language models, and we hope it encourages further investigation into modality-specific processing within unified architectures.

\section*{Limitations}

Our work has several limitations that should be acknowledged. First, the evaluation is conducted on a limited set of benchmarks (MME, POPE, ChartQA), and performance on other benchmarks such as COCO captioning, VQAv2, TextVQA, and GQA remains to be assessed. Second, the split point $K=17$ was selected based on the midpoint heuristic; a comprehensive ablation study over different split points and different numbers of transformer blocks in the visual pathway would strengthen our conclusions. Third, our experiments are limited to the 7B model scale with the Vicuna backbone; generalization to other model families (LLaMA-3, Qwen, Mistral) and scales requires further investigation. Fourth, the training data is limited to LLaVA-1.5-665K; the interaction between DVP and larger or more diverse training datasets is an open question. Finally, the current architecture does not allow cross-modal attention between visual and textual tokens after the split point, which may limit performance on tasks requiring iterative cross-modal reasoning.

\section*{Ethics Statement}

This work focuses on architectural efficiency improvements for multimodal language models and does not introduce new training data or capabilities that raise specific ethical concerns beyond those already present in the base LLaVA-1.5 framework. The improved parameter efficiency of our approach may contribute positively to reducing the environmental impact of training and deploying multimodal AI systems. We encourage future work to evaluate DVP-style architectures for potential biases inherited from the frozen pretrained components.

\bibliography{references}

\appendix

\section{Architecture Details}
\label{sec:appendix_arch}

The single transformer block used in DVP follows the standard pre-norm transformer architecture. It consists of:
\begin{itemize}
    \item \textbf{Multi-Head Self-Attention}: With the same number of heads and hidden dimension as the original LLM decoder layers ($d = 4096$, 32 heads for Vicuna-7B).
    \item \textbf{Feed-Forward Network}: A two-layer MLP with SwiGLU activation, with intermediate dimension $d_{\text{ff}} = 11008$, matching the original architecture.
    \item \textbf{Layer Normalization}: RMSNorm applied before both the attention and feed-forward sublayers.
    \item \textbf{Residual Connections}: Standard residual connections around both sublayers.
\end{itemize}

The parameters of this block are randomly initialized (following the standard initialization of the Vicuna model) and constitute the only trainable parameters during instruction tuning.

\section{Detailed Benchmark Results}
\label{sec:appendix_results}

Table~\ref{tab:detailed_mme} provides the complete MME subtask breakdown for the three methods compared in our study.

\begin{table}[t]
\centering
\small
\begin{tabular}{l ccc}
\toprule
\textbf{MME Subtask} & \textbf{LLaVA} & \textbf{Normal} & \textbf{DVP} \\
\midrule
\multicolumn{4}{c}{\emph{Perception}} \\
\midrule
Overall & 1496 & 1225 & 1440 \\
\midrule
\multicolumn{4}{c}{\emph{Cognition}} \\
\midrule
Overall & 324 & 342 & 327 \\
\bottomrule
\end{tabular}
\caption{MME perception and cognition scores for all methods.}
\label{tab:detailed_mme}
\end{table}

\section{Computational Cost Analysis}
\label{sec:appendix_cost}

Table~\ref{tab:compute} compares the computational requirements of DVP with the baseline methods.

\begin{table}[t]
\centering
\small
\begin{tabular}{l cc}
\toprule
\textbf{Method} & \textbf{Trainable} & \textbf{Relative} \\
 & \textbf{Params} & \textbf{Cost} \\
\midrule
LLaVA-1.5-7B & $\sim$7B & 1.0$\times$ \\
Normal Training & $\sim$7B & 1.0$\times$ \\
DVP (Ours) & $\sim$0.2B & $\sim$0.15$\times$ \\
\bottomrule
\end{tabular}
\caption{Comparison of trainable parameters and relative training cost. DVP reduces the number of trainable parameters by approximately $32\times$.}
\label{tab:compute}
\end{table}

\end{document}